\documentclass{article}

\usepackage{arxiv}
\usepackage[utf8]{inputenc} 
\usepackage[T1]{fontenc}   
\usepackage{hyperref}      
\usepackage{url}       
\usepackage{booktabs}   
\usepackage{amsfonts}     
\usepackage{nicefrac}    
\usepackage{microtype}
\usepackage{lipsum}
\usepackage{graphicx}
\usepackage[authoryear]{natbib}
\usepackage{amssymb}
\usepackage[figuresright]{rotating}
\usepackage{tikz}
\usepackage{amsmath}
\usepackage{algorithm}
\usepackage[noend]{algpseudocode}
\usepackage{stfloats}

\title{SubTGraph: Large-Scale Subterranean Environment Synthesis with Controllable Topological Variability for Robotic Autonomy Validation}

\author{
  Fernando Labra Caso%
  \thanks{This work has been partially funded by the European Union’s Horizon Europe Research and Innovation Program, under the Grant Agreement No. 101119774 SPEAR.} \hspace{0.1mm}
  \thanks{This work has been partially funded by the European Union’s Horizon Europe Research and Innovation Program, under the Grant Agreement No.101138451 PERSEPHONE.} \\
  Robotics \& AI\\
  Luleå University of Technology \\
  \texttt{fernando.labra.caso@ltu.se} \\
   \And
  Akshit Saradagi \\
  Robotics \& AI\\
  Luleå University of Technology \\
  \texttt{akshit.saradagi@ltu.se} \\
   \And
  Scott Fredriksson \\
  Robotics \& AI\\
  Luleå University of Technology \\
  \texttt{scott.fredriksson@ltu.se} \\
   \And
  Samuel Nordström \\
  Robotics \& AI\\
  Luleå University of Technology \\
   \And
  Anton Koval \\
  Robotics \& AI\\
  Luleå University of Technology \\
  \texttt{anton.koval@ltu.se} \\
   \And
  George Nikolakopoulos \\
  Robotics \& AI\\
  Luleå University of Technology \\
  \texttt{george.nikolakopoulos@ltu.se} \\
}

\begin{document}

\maketitle

\begin{abstract}
Subterranean (SubT) environments have been a frontier for autonomous robotics, driven by the push for automation of mining operations and the interest in planetary exploration (Martian Lava Tubes). Due to the challenges involved in accessing real SubT environments, rigorous hardening of autonomy stacks in realistic simulation environments is critical. This article fills a well-known gap, which relates to the unavailability of a large-scale simulation-based benchmarking infrastructure for rigorous statistical evaluation of robotic autonomy, due to which it is common for SubT research articles to present validation results in a few environments at best. 

This article presents SubTGraph, a novel framework for rapid synthesis of multi-level SubT environments with high variability, incorporating user specifications related to topology, dimensionality, textures, etc., to generate distinct environments such as operational mines, natural caves and lava tubes. SubTGraph builds a cost matrix from user-specified structural constraints to guide the classical Dijkstra algorithm to procedurally generate SubT worlds utilizing topometric tiles from the DARPA World Generator. Three robotics case-studies are investigated to demonstrate the utility of SubTGraph for rigorous validation of different layers in the robotic autonomy stack. Structural semantic segmentation is validated against topometric ground truths, multi-agent path planning is widely tested for identification of patterns and trends in the algorithm behavior and LIO SLAM is stress-tested in challenging subterranean sections to identify failure cases. The SubTGraph world creation codebase is open-sourced (\href{https://github.com/LTU-RAI/SubTGraph.git}{https://github.com/LTU-RAI/SubTGraph.git}) along with a database consisting of 150 highly variable underground worlds. 
\end{abstract}

\twocolumn

\begin{figure*}[t]
    \centering
    \includegraphics[width=\textwidth]{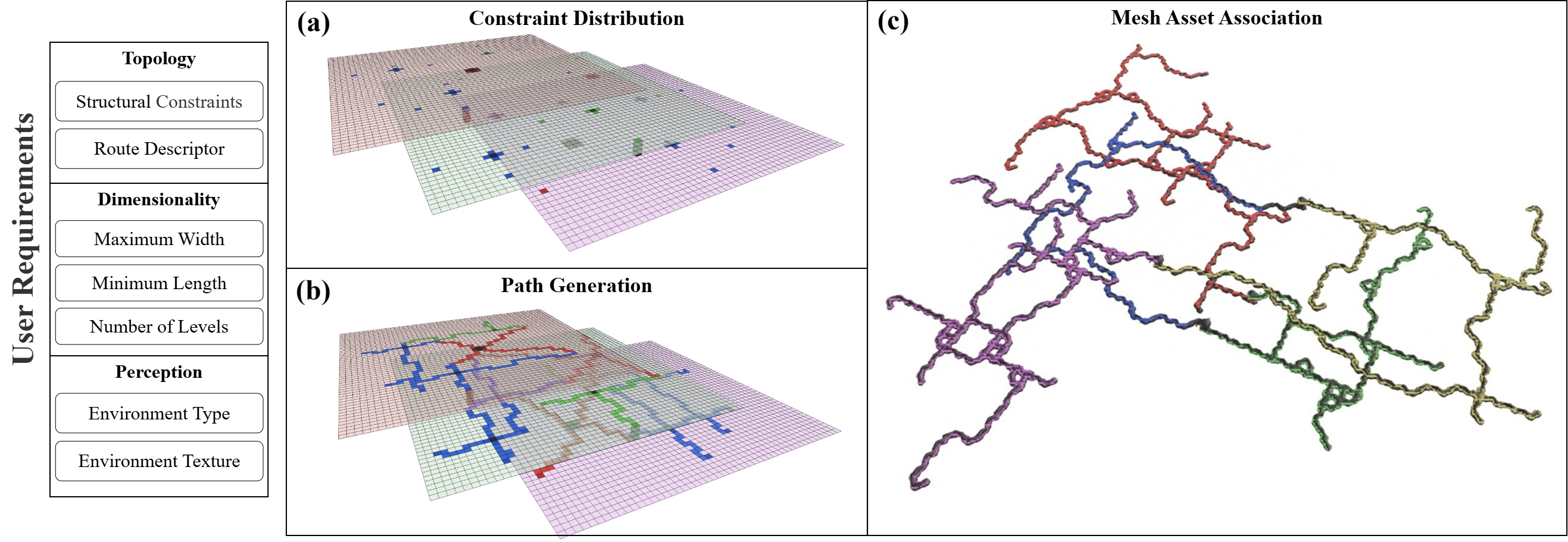} \vspace{-3mm}
    \caption{SubTGraph Conceptual Diagram. User requirements specify control over the topology, dimensionality and perception of the generated environment. These values are used to distribute a set of structural constraints and objective nodes. Circuits are created by defining Dijkstra cost matrices that follow route descriptors, increasing the topological variability. 3D meshes are associated per level to produce a physics simulator agnostic validation world.} \vspace{-3mm}
    \label{fig:figure1}
\end{figure*} 

\section{Introduction} 

\noindent Simulations are an essential tool in the verification process of robotic algorithms previous to field deployment \citep{collins2021review}. The growing need for simulated environments in different research areas contrasts with the short availability of environment creation tools \citep{koval2020subterranean}. Algorithms including Simultaneous Localization and Mapping (SLAM) or Lidar Inertial Odometry (LIO) require the usage of point clouds through sensor readings. However, the deployment of such solutions in subterranean environments presents a set of challenges with respect to data collection and testing, e.g. time allocation or data quality and variance \citep{pauly2004uncertainty}, \citep{javaheri2017subjective}. Furthermore, the accessibility of research groups to subterranean areas is not only constrained by geographical availability, but also security requirements and partnership agreements. These conditions slow the development of subterranean robotics research.
Triangular meshes provide a sufficient approximation for algorithms requiring low-detail subterranean environments. The classical method for mesh generation is point cloud surface reconstruction \citep{huang2024surface}. This method is applied with classical triangulation models, such as Poison Surface Reconstruction \citep{kazhdan2013screened}, or Deep Learning models e.g. IMLSNet \citep{liu2021deep}, which apply non-linear transforms to generate a denser 3-dimensional representation to be triangulated.
Reconstructed point clouds correctly represent the shape of an underground area, but suffer from noisy data and produce highly complex meshes. Simple geometrical environments don't achieve realistic requirements but excel in parallel optimization tasks. This gap in the capabilities and size of the environments is completed with the introduction of virtual assets from the DARPA SubT challenge \citep{ackerman2022robots}. In this challenge, teams compete in identifying sets of objects when exploring subterranean worlds both in real deployment and virtual simulations under a preset duration. DARPA SubT provides one of the first subterranean world generators, acting as benchmark for their associated navigation and identification challenge.

Subterranean navigation and mapping is key in robotics research \citep{thrun1996integrating}. The characteristics of subterranean areas for robotic tasks in terms of texture, size, heterogeneity and variance are difficult to portray in simulation but necessary to achieve realistic field transfers \citep{koval2020subterranean}.
This work presents SubTGraph, a procedural subterranean world generator that allows topological control. This tool serves the statistical evaluation of robotic autonomy methods in cave environments, expediting simulated validation before deployment by providing a virtual testing platform. The usage of structural constraints in the form of tile connections (junctions, loops, intersections) provides the user with the ability to parametrize the topology of the underground environment. The virtual worlds are represented with occupancy matrices, facilitating mesh storage and reproduction. A dataset of 150 cave worlds with different topologies, structural constraints and parameters is released to satisfy the needs of a benchmarking platform in the underground robotic autonomy research. Figure \ref{fig:figure1} displays a conceptual diagram of the components and generation process of the SubTGraph creation tool. \vspace{-3mm}

\section{Problem Overview}

\subsection{Subterranean Unpredictability} 

\noindent Subterranean environments present challenges outside any standard robotic autonomy stack. Some of these challenges are directly related to communication, localization, exploration, navigation and mapping. Caves develop tunnel obstructions that change their topology either instantaneously or over time. Such obstructions, cluttered tunnels and complex labyrinth-like structures produce great disturbances in the signals received and sent by the robot \citep{walsh2018communications}. Physical obstacles, mine clutter, complex topologies, uneven ground and low light conditions increase difficulty in the deployment, navigation and exploration of robots \citep{silver2006topological} \citep{bakambu2007autonomous}. Long, parallel and continuous tunnels with high-degree visual aliasing present challenges for camera and lidar localization methods by producing large uncertainties in the discovery of unique features \citep{jacobson2021localizes}. Absence of common navigation sensors such as GPS or GNSS presents an added challenge in the localization stack \citep{losch2018design}. 

\subsection{DARPA SubT Challenge}

\noindent The DARPA subterranean challenge pursued Virtual and Systems competitions for respectively best results on simulated underground worlds and real-life operated mines. The outcome of this challenge is an ecosystem of robotic methods and datasets for exploration, navigation, localization and mapping. However, no simulation worlds for statistical validation were released from such challenge; leaving the virtual competition environments and the DARPA online world creator as the only means to generate testing platforms. The teams where presented with a set of virtual worlds created from dense point clouds and individual tile assets along with ground truth maps which were manually surveyed. 

Virtual competition members express how synthetic worlds provide better results than manually scanned ones. Their solutions were tested primarily on the tile-based worlds but still yielded competitive results on the real sets \citep{chung2023into}. Limitations of the robotic solutions when facing crossroads and long narrow corridors are also mentioned as challenges of the simulation testing \citep{bayer2023autonomous}, stressing the need for extended validation of software solutions previous to deployment. Along this line, virtual teams remarked the need for a simulation platform that can produce and represent a wide variety of real-world caves based on an algorithmic and controllable method with topological and terrain parameters: e.g. number of cycles, frequency of intersections or traversability ratios \citep{chung2023into}. The robotics community suggests such platform for safe, fast and low-cost virtual proving grounds. Large amounts of training data for verifiable behaviors and repeatability for corner cases together with risk-free controllable experiments democratizes the simulation-in-robotics effort and fosters open-source tools and validation metrics \citep{choi2021use}. 

\subsection{SubTGraph Creation Tool}

\noindent In order to tackle the aforementioned challenges and develop a robust and accountable robotics method that is extensively validated on general and corner cases; the SubTGraph creation tool provides a fast, explainable, computationally low method to generate subterranean virtual worlds on user-defined geometrical, topological, visual and textural parameters. Geometrical features include the minimum size of the environment along with the maximum size of the tunnels. Topological features are the core of this tool and utilize structural constraints as number of junctions, loops and intersections used in the creation of the subterranean world. The visual and textural perception of the environment can be altered by providing the desired texture during generation. Moreover, rock obstacles are spawned along the environment according to the user specifications. 

This work aims to present three different scenarios for the current development of robotic methods: (1) Structured, (2) Unstructured and (3) Lava tube virtual worlds. The structured and unstructured environments represent respectively operational mines and natural caves whereas the lava tubes mimic the underground topologies found in volcanic areas of Northern Europe but also present below the martian surfaces. Challenges arising from highly variant, niche subterranean environments produce unexpected results during field tests and require large amounts of effort and time to provide a functioning deployable method. Statistical evaluation is presented in this work as the means for validation under different pretexts, encapsulating in every virtual environment the necessary and abstract features that will produce a correct generalized method. These features allow the user to benchmark a method based on the chosen specifications and stress-test general and corner cases for future improvement. In this work, several use cases are tested including localization, mapping and path planning methods to demonstrate the utility of this tool.

The transitioning of autonomous robotic methods to real operational mining environments highlights the importance of such validation tools. This research remarks that accountability and explainability can be reasonably achieved on simulation without great costs, allowing the development of robotics methods under low risk workspaces and expediting the usage of virtual twins during real-life deployments. \vspace{-4mm}

\subsection{Problem definition}

\noindent Subterranean environments are difficult to model due to topological, textural and clutter complexity. The DARPA SubT challenge presented the first set of triangular meshes with variety in textures and shapes for individual topometric components such as intersections, junctions, corners, pathways, shafts and deadends. These assets were released with a manual world builder and a simulation generator that aligned with the challenge objectives, allowing the validation of robotic autonomy stacks in the participating teams. However, little controllability over the complexity, the dimensionality or the texture of the environment was offered; rendering both builder and simulator obsolete for general statistical validation.

Currently, subterranean robotics research teams utilize these tools to generate a handful of environments in which to test their algorithms. This step is time consuming for both manually created and generated topologies, as the representation of the virtual world is platform dependent for the standard libraries and simulator tools at the time of the challenge release. The robotics community expresses a need for a controllable, ease-of-use platform that includes the individual requirements for statistical validation of autonomy stacks in varying environments.

\begin{figure*}[t]
    \centering
    \includegraphics[width=\textwidth]{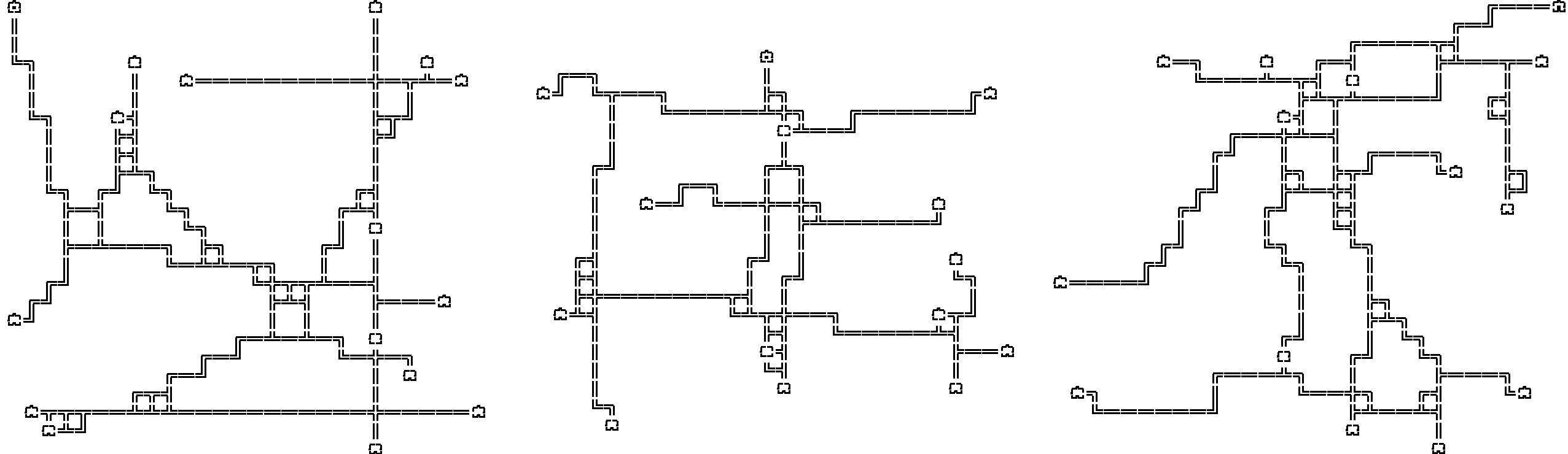} \vspace{-3mm}
    \caption{The algorithm creates 2-dimensional topologies for each specified level by utilizing the route descriptors to generate straight, wide curved or wave-like circuits. This figure shows three example topologies; each element is translated onto mesh objects and vertical/horizontal offsets are applied.} \vspace{-3mm}
    \label{fig:figure2} 
\end{figure*}
 
\section{Methodology} 

\noindent The SubTGraph world generation procedure discretizes the underground topology into grid components. In order to facilitate the mathematical relationship between them, this method assumes a graph-based approach. The connectivity of the nodes defines the topometric units, e.g. intersections, loops; described by the neighboring occupancies in the grid. The topology is obtained as the accumulation of estimated shortest paths between constraint and objective nodes at each level. These paths are mapped into grid positions and provide the connectivity between the nodes of the graph.

In order to achieve this, an initial node set is distributed along the grid, with the occupied positions becoming constraint nodes, as seen in Fig. \ref{fig:figure1}. The node sets are obtained from user requirements to satisfy at least the number of loops, intersections and junctions in the grid. Every constraint node for each node set is associated with an objective node in a randomly selected position, defining a potential path. The path is constructed as a linear segment and distorted by applying a route descriptor (i.e. harmonic signal). Discretizing this path yields an initial cost matrix between constraint and objective nodes. The cost matrix is not equivalent to an occupancy matrix, as only a reduced number of points is mapped to guide the Dijkstra process, while increasing randomness and stochasticity of the generation procedure. Dijkstra's algorithm provides the path from the constraint to the objective node and their continuous accumulation produces the topological description of a 2-dimensional level. Moreover, the interconnection of the occupancies through vertical tiles, by selecting an objective node that acts as shaft at each level, allows the creation of multi-level topologies.

The directional connectivity of each node defines their topometric tile as junctions, intersections, etc; thereafter the process selects and merges the associated mesh assets to produce a virtual world. Visual textures are applied on the meshes, resulting in a realistic 3-dimensional simulated environment. With this procedure (1) the dimensionality is controlled by both the grid size and the position of the objective nodes, (2) the topology is directed through the route descriptor and (3) the textures applied on the mesh provide varying visual characteristics that differentiate the environment types.

\begin{algorithm}[h!]
\caption{SubTGraph Generation Algorithm}
\small
\begin{algorithmic}[1]
    \State Initialize  $\{l,m,n \in \mathbb{N}, \ W \in \{0,1\}^{l \times m \times n} \}$
    \State \textbf{a) Structural Definition ------------------------------------------}
    \State Initialize  $\{S\in \mathbb{N}^{N_C \times 2},\hat{C}\in \mathbb{N}^{N_C \times 8 \times 2}, \ \hat{O} \in \mathbb{N}^{N_C \times 8 \times 2}\}$
    \For{$[k=0; \ k < N_C; \ k = k+1]$} \scriptsize \textbf{\textit{\text{\% For Every Requirement}}} \small
        \State \scriptsize \textbf{\textit{\text{\% Starting Point}}} \small
        \State $S[k]:=(x_s,y_s)=( rand(1,n), rand(1,m))$ 
        \State $\phi = rand(\{J,L,I\}); \ \iota = rand(0,3)$
        \State $\hat{C}[k] := f_c(S[k], \Upsilon^\phi ,\rho^{\phi,\iota})$ \scriptsize \textbf{\textit{\text{\% Compute Constraint Set}}} \small
    \EndFor
    \ForAll{$ [C \in \hat{C} ]$}  \scriptsize \textbf{\textit{\text{\% For Every Constraint Set}}} \small
        \State $k_{out}=pos(C, \hat{C})$
        \ForAll{$ [(x_c,y_c) = c \in C ]$}  \scriptsize \textit{\textbf{\text{\% For Every Constraint}}} \small
            \State $k_{in}=pos(c, C) $
            \State $\delta = rand(\{-1,1\}); \ \mu = rand(-2,2)$ 
            \If{$\beta_H$} \scriptsize \textbf{\textit{\text{\% Horizontal Translation}}} \small
                \State $\Delta_x =\delta * rand(2,n); \ \Delta_y = \mu$
            \EndIf
            \If{$\beta_V$} \scriptsize \textbf{\textit{\text{\% Vertical Translation}}} \small
                \State $\Delta_y=\delta*rand(2,m); \ \Delta_x=\mu$
            \EndIf
            \State $\hat{O}[k_{out}][k_{in}] := (x_c + \Delta_x, y_c + \Delta_y)$ \scriptsize \textbf{\textit{\text{\% Objective Point}}} \small
        \EndFor
    \EndFor
    \State \textbf{b) Path Generation ----------------------------------------------}
        \State \scriptsize \textbf{\textit{\text{\% For Every Constraint and Objective Sets}}} \small
        \ForAll{$[C \in \hat{C} , \ O\in \hat{O} ]$}  
            \State Initialize $\{\Phi \in \mathbb{N}, \ W_{cost} \in \mathbb{R}^{m\times n} \}$
            \ForAll{$[(x_c,y_c) = c \in C , \ (x_o,y_o) = o \in O]$}
                \State $\vec{dx}=\langle x_o-x_c, \ 0 \rangle; \ \vec{dy}=\langle 0, \ y_o-y_c \rangle$
                \State  \scriptsize \textbf{\textit{\text{\% Linear Segment Interpolation}}} \small
                \State $S^P=\Phi \cdot L = \Phi \cdot ||\vec{dx}+\vec{dy}||_2$ \scriptsize \textbf{\textit{\text{\% Number of Samples}}} \small
                \For{$[s=0; \ s < S^P; \ s = s + 1]$}
                    \State $p=s/S_P$ \scriptsize \textbf{\textit{\text{\% Distance Ratio}}} \small
                    \State $(x_p, y_p) = (f_x(p), f_y(p))$  \scriptsize \textbf{\textit{\text{\% Sample Point}}} \small
                    \State $o = f_o(p)$ \scriptsize \textbf{\textit{\text{\% Route Descriptor Offset}}} \small
                    \State \scriptsize \textbf{\textit{\text{\% Discrete Path Guide Point}}} \small
                    \State $x_w= max(\lfloor x_p + (o \cdot \vec{dy}^{\perp}) \rfloor, n) $
                    \State $y_w= max(\lfloor y_p + (o \cdot \vec{dx}^{\perp}) \rfloor, m)$  
                    \State $W_{cost}(x_w, y_w) = 1$ \scriptsize \textbf{\textit{\text{\% Update Cost Matrix}}} \small
                \EndFor
                \State $O^{map}_{(c, o)}=dijkstra(W_{cost},c,o)$ 
                \State \scriptsize \textbf{\textit{\text{\% Compute Shortest Path}}} \small
                \State $W = W \cup O^{map}_{(c, o)}$ \scriptsize \textbf{\textit{\text{\% Update Occupancy Matrix}}} \small 
        \EndFor 
    \EndFor
    \State \textbf{c) Mesh Association ----------------------------------------------}
    \State Initialize $\{ \hat{W}_{asset} \in \mathbb{R}^{m\times n} \}$
    \State $p_{start} = \hat{O}[0][0] = (x_o, y_o); \ p_{prev} = \emptyset $ \scriptsize \textbf{\textit{\text{\% Start Conditions}}} \small
    \State $f_m(W, p_{start}, p_{prev})$ \scriptsize \textbf{\textit{\text{\% Recursive Mesh Instantiation}}} \small 
\end{algorithmic}
\label{alg:algo1}
\end{algorithm}
\normalsize

\subsection{Subterranean World Discretization}

\noindent A matrix $W_{occupancy} \in \{0,1\}^{l \times m \times n}$ defines the 3-dimensional space that the subterranean environment occupies with $l$ levels, $m$ rows and $n$ columns. A horizontal slice of this matrix represents a 2-dimensional map of one subterranean level with each grid element an intersection, junction, corner, corridor or deadend connection based on the neighboring occupancy. For example, a junction connection extends towards three distinct directions from any grid position.

\subsection{Structural Definition}

\noindent The user structural requirements are utilized to spawn the constraint node sets $C$ at different starting points and levels of the grid. These node sets represent the required occupancies for the algorithm to identify the respective topometric units. For each constraint node in the constraint sets a random objective node $o$ is obtained along guided areas, yielding objective node sets $O$. The constraint node sets are defined as per Eq. \ref{eq1}. \vspace{-3mm}

\small

\[
\begin{aligned}
    \vec{\upsilon_r}=(1,0); \quad \vec{\upsilon_l}=(-1,0); \quad \vec{\upsilon_d}=(0,1); \quad \vec{\upsilon_u}=(0,-1)
\end{aligned}
\] \vspace{-6mm}

\begin{equation}\label{eq1}
    \begin{aligned}
        \Upsilon^J &= \{ \vec{\upsilon_u}, \vec{\upsilon_d}, \vec{\upsilon_r} \} \\
        \Upsilon^I &= \Upsilon^J \cup \{ \vec{\upsilon_l}\} \\
        \Upsilon^L &= \Upsilon^I \cup \{ \vec{\upsilon_u} + \vec{\upsilon_r},  
        \vec{\upsilon_u} + \vec{\upsilon_l}, \vec{\upsilon_d} + \vec{\upsilon_r}, 
        \vec{\upsilon_d} + \vec{\upsilon_l}\}
    \end{aligned}
\end{equation} 

\normalsize

\subsubsection{Constraint Node Distribution}

\noindent Let the random starting position within a 2-dimensional level be denoted as \( s=(x_s, \ y_s) \),  the node sets are applied to obtain the constraint nodes as per Eq. \ref{eq4}. For this matter, the operator $\oplus$ is defined in Eq. \ref{eq2} as the translation of a point in the direction of a set of vectors and Eq. \ref{eq3} introduces a rotation effector that allows all permutations of the node sets $\Upsilon$, as they represent a fixed direction for each unit. \vspace{-3mm}

\small

\begin{equation}\label{eq2}
    \begin{aligned}
        a = (x_a, y_a); \ B = \{\vec{b_1}, \vec{b_2}\}; \ a \oplus B = \{ a+\vec{b_1}, a + \vec{b_2}\}
    \end{aligned}
\end{equation} \vspace{-8mm}

\begin{equation}\label{eq3}
    \begin{aligned}
        \rho^{\phi,\iota} =
            \begin{cases}
             1  & \text{if } \phi \in \{L,I\} \\
            (\iota \cdot \pi)/2; \quad \iota \in [0,3]  &  otherwise
            \end{cases}
    \end{aligned}
\end{equation} \vspace{-6mm}

\begin{equation}\label{eq4}
    \begin{aligned}
        f_c(s,\Upsilon^\phi, \rho^{\phi,\iota}) = (s \oplus \Upsilon^\phi) \cdot \rho^{\phi,i}, \quad \phi \in \{J,L,I\}
    \end{aligned}
\end{equation} 

\normalsize

\subsubsection{Objective Node Distribution}

\noindent A random objective node $o \in O$ is obtained for every constraint node in the selected set $c \in C$. The objective position \( (x_o, y_o) \) is computed by horizontally ($\beta_H$) or vertically ($\beta_V$) translating the constraint position within a specified range as defined in equations Eq. \ref{eq5} and Eq. \ref{eq6}. \vspace{-6mm}

\[
    \begin{aligned}
         \delta \in \{-1,1\},\quad t_x \in [2, n],\quad t_y \in [2, m], \quad \mu \in [-2, 2]
    \end{aligned}
\] \vspace{-10mm}

\[
    \begin{aligned}
         \beta_V \rightarrow y_c = y_s \land \neg (x_c = x_s) \quad  \beta_H \rightarrow x_c = x_s \land \neg (y_c = y_s)
    \end{aligned}
\] \vspace{-10mm}

\begin{equation}\label{eq5}
    \begin{aligned}
        \Delta_x =
        \begin{cases}
         \delta * t_x  & \text{if } \beta_H \\
         \mu  &  \text{if } \beta_V
        \end{cases} \quad
        \Delta_y =
        \begin{cases}
         \mu  & \text{if } \beta_H \\
         \delta * t_y  &  \text{if } \beta_V
        \end{cases}
    \end{aligned}
\end{equation} \vspace{-8mm}

\begin{equation}\label{eq6}
    o = (x_o, y_o) = (x_c +\Delta_x, \ y_c + \Delta y)
\end{equation}

\subsection{Path Generation}

\noindent After distributing all constraint and objective node sets, the algorithm computes a cost matrix by calculating and distorting a linear segment between each constraint and objective nodes. Thereafter, a subset of segment points is discretized and mapped onto the cost matrix to guide the Dijkstra path discovery. The distortion of the linear segment is achieved by applying a route descriptor that represents the shape tendencies of the desired subterranean world. The route descriptors are built as varying harmonic signals. This work has explored linear, parabolic and sine functions as respectively $\eta\in\{0,1,2\}$ number of harmonics: \vspace{-1mm}

\begin{enumerate}
    \item Linear descriptor for straight circuits.
    \item Parabolic descriptor for wide curved circuits.
    \item Sine descriptor for wave-like circuits.
\end{enumerate}

\subsubsection{Constraint Segment Construction}

\noindent  Let the start point of the linear segment be \( (x_c, y_c) \) and the end point be \( (x_o, y_o) \). Let \( \vec{dx} \) and \( \vec{dy} \) denote the components of the direction vector as per Eq. \ref{eq7}. The discrete sample points $S^P$ along the line are computed proportional to its Euclidean length \( L \) by a factor \( \Phi \) following Eq. \ref{eq8}. \vspace{-4mm}

\begin{equation}\label{eq7}
    \vec{dx}=\langle x_o-x_c, \ 0 \rangle; \quad \vec{dy}=\langle 0, \ y_o-y_c \rangle
\end{equation} \vspace{-6mm}

\begin{equation}\label{eq8}
    S^P = \Phi \cdot L = \Phi \cdot ||\vec{dx}+\vec{dy}||_2
\end{equation}

\noindent The linear segment is parameterized as per Eq. \ref{eq10} with the translation of the starting point by a distance factor of the total length. This formulation allows the discretization of the segment by using the normalized value $p_i$ as per Eq. \ref{eq9}. \vspace{-5mm}

\begin{equation}\label{eq9}
p_i = \frac{s_i}{S^P}, \quad s_i \in [0, S^P]
\end{equation} \vspace{-8mm}

\begin{equation}\label{eq10}
   x_p = f_x(p) = x_c + \vec{dx} \cdot p; \quad y_p = f_y(p) = y_c + \vec{dy} \cdot p
\end{equation}

\begin{figure*}[t!]
    \centering
    \includegraphics[width=\linewidth]{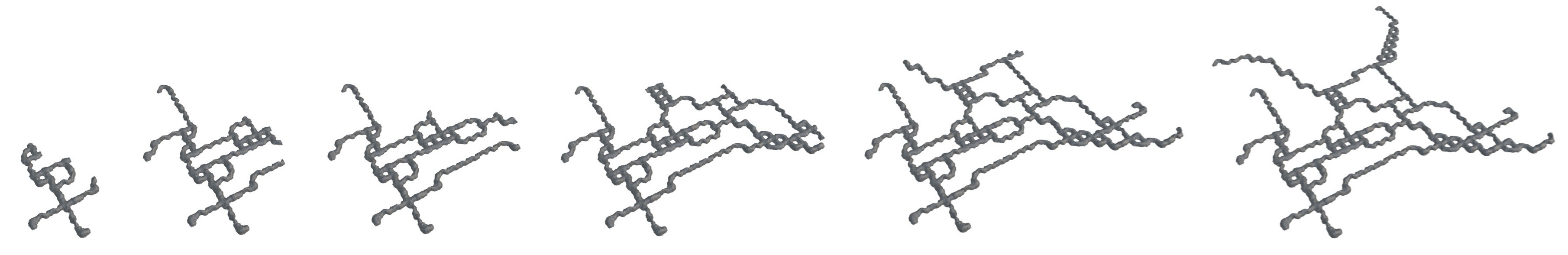}
    \caption{This figure displays the recursive generation process of the mesh object. Once the constraint distribution is complete and the Dijkstra occupancy matrix has been obtained, the mesh asset association initiates the generation process at the starting node and recursively adds mesh components through each possible direction until a circuit deadend is found.} \vspace{-3mm}
    \label{fig:figure3}
\end{figure*} 

\subsubsection{Constraint Segment Distortion}

\noindent The distortion is calculated as per Eq. \ref{eq11} based on the user defined harmonic $\eta$. The amplitude \( A \) of the wave is computed from the dimensions of the grid. \vspace{-6mm}

\begin{equation}\label{eq11}
    f_o(p) = A \cdot \sin(\nu \cdot p) = \frac{1}{2 \cdot \min(m, n)} \cdot \sin(\eta \cdot \pi \cdot p)
\end{equation}

\noindent The segment points are translated perpendicularly with the direction vectors in Eq. \ref{eq12}. The resulting discretized and distorted points $(x_w,y_w)$ are obtained as per Eq. \ref{eq13}. \vspace{-3mm}

\begin{equation}\label{eq12}
    \vec{dx}^{\perp} = -dy \cdot L^{-1}, \quad \vec{dy}^{\perp} = dx \cdot L^{-1}    
\end{equation} \vspace{-10mm}

\begin{equation}\label{eq13}
    \begin{matrix} 
        x_w = g_x(p) = max( \lfloor f_x(p) + f_o(p) \cdot \vec{dx}^{\perp} \rfloor, n) \\ 
        y_w = g_y(p) = max(\lfloor f_y(p) + f_o(p) \cdot \vec{dy}^{\perp} \rfloor, m) 
    \end{matrix}
\end{equation}

\begin{figure}[b!]
    \centering
    \includegraphics[width=\linewidth]{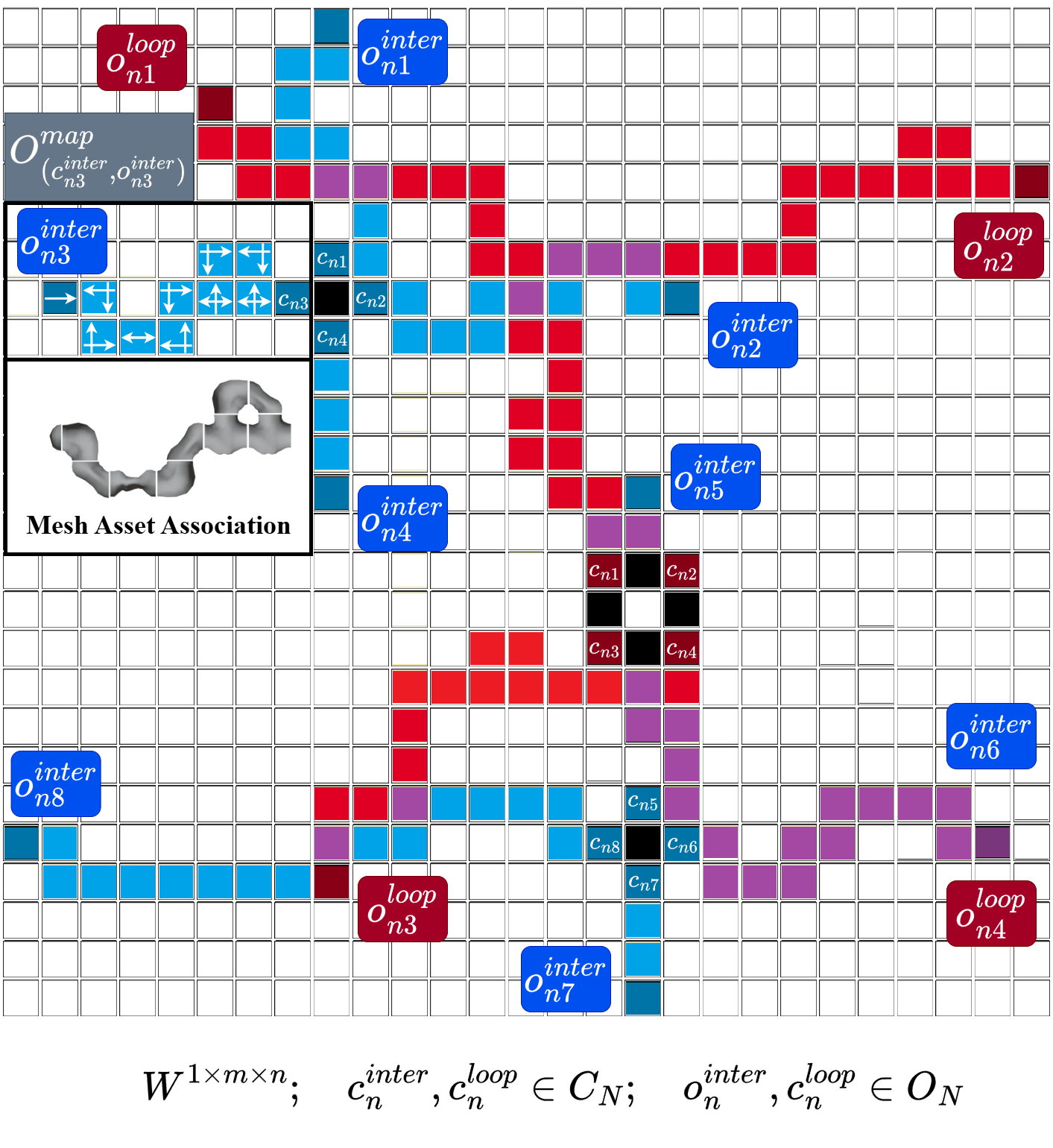} \vspace{-6mm}
    \caption{Single-level Occupancy Matrix. Two intersections and one loop (blue/red respectively) are utilized as structural constraints.An example of the mesh asset association process is shown for the occupancy map generated between constraint and objective nodes $(c_{n3}^{inter},o_{n3}^{inter})$.} \vspace{-6mm}
    \label{fig:figure4}
\end{figure}

\subsubsection{Dijkstra Path Estimation}

\noindent The cost matrix $W_{cost} \in \mathbb{R}^{m \times n}$ for each constraint and objective node pairs is initialized with prohibitive weights except for the elements of the discretized distorted segment as per Eq. \ref{eq14} and Eq. \ref{eq15}. This guides the Dijkstra process to follow the shape tendencies defined with the route descriptor. \vspace{-5mm}

\small

\begin{equation}\label{eq14}
W_{cost}(g_x(p_i), g_y(p_i)) = 1; \quad \forall p_i = \frac{s_i}{S^P}, s_i \in [0, S^P]    
\end{equation} \vspace{-12mm}

\begin{equation}\label{eq15}
W_{cost}(u, v) = \infty; \quad \forall u \forall v  \in V - \{g_x(p_i), g_y(p_i)\} 
\end{equation} \vspace{-5mm}

\noindent Dijkstra's shortest path algorithm is computed for each cost matrix. The occupancy matrix $W_{occupancy}[k] \in \{0,1\}^{1 \times m \times n}$ of the current level is updated with the union of each resulting occupancy map $O^{map}_{(c, o)}$ as per Eq. \ref{eq16} until all constraint and objective pairs have been resolved. \vspace{-3mm}

\begin{equation}\label{eq16}
    W_{occupancy}[k] \cup O^{map}_{(c, o)}; \quad \forall k \in [0,l], \forall c \in C, \ \forall o \in O
\end{equation}

\noindent Figure \ref{fig:figure4} displays a single level example occupancy matrix with one loop, and two intersection node sets. Each constraint type and estimated path is associated with a color: red for loop constraints and blue for intersection constraints. 

\subsection{Mesh Association}

\noindent At each level $l$ of the occupancy matrix, the mesh building component recursively identifies deadends, pathways, corners, junctions and intersections based on the neighboring nodes as per Eq. \ref{eq17}. Figure \ref{fig:figure4} shows the process followed during one recursive call in the generated path between the constraint and objective node pair $(c_{3},o_{3})$. The number of connections defines the topometric type and their direction indicates the necessary rotation during mesh instantiation. The virtual world $\hat{W}$ is updated with randomly selected mesh surfaces for each topometric type from a user defined list of available DARPA SubT assets $M^\phi \ \forall\phi \in \{D,C,P,J,I\}$. Afterwards, the generation process applies an offset to adjust vertically and horizontally every new mesh connection. \vspace{-4mm}

\begin{equation}\label{eq17}
    \alpha(W_{occupancy},p \cdot \rho^{\phi,\iota}) \rightarrow W[p_x][p_y] = 1
\end{equation} \vspace{-10mm}

\[
    \begin{aligned}
        \beta_D &\rightarrow \alpha(W, \ p + \vec{\upsilon_r}) & &\\
        \beta_P &\rightarrow \beta_D \land \alpha(W, \ p + \vec{\upsilon_l}); & \beta_C &\rightarrow \beta_D \land \alpha(W, \ p + \vec{\upsilon_d}) \\
        \beta_J &\rightarrow \beta_C \land \alpha(W, \ p + \vec{\upsilon_u}); & \beta_I &\rightarrow \beta_J \land \alpha(W, \ p + \vec{\upsilon_l})
    \end{aligned}
\] \vspace{-2mm}

\noindent This recursive process is described in Algorithm \ref{alg:algo2} and shown in Figure \ref{fig:figure3}. After updating the virtual world, one or more vacant positions according to the occupancy and the topometric type are selected as per Eq. \ref{eq18} and the process is recursively repeated until the new position has already been filled or a dead end is reached. \vspace{-4mm}

\begin{equation}\label{eq18}
     \begin{aligned}
        \gamma(p, \vec{\upsilon}_{start}) =
        \begin{cases}
            p^* = p + \vec{\upsilon} &\text{if } \hat{W}[p^*_x][p^*_y] = \emptyset \\
            \emptyset & \text{otherwise}
        \end{cases}
        \\ \forall \vec{\upsilon} \in \{\vec{\upsilon_u}, \vec{\upsilon_d}, \vec{\upsilon_l}, \vec{\upsilon_r}\} - \vec{\upsilon_{start}}
    \end{aligned}   
\end{equation} 

\begin{algorithm}[h!]
\caption{Mesh Association Recursive Algorithm}
\small
\begin{algorithmic}[1]
    \If{$\hat{W}[p_{start_x}][p_{start_y}] = \emptyset$} \scriptsize \textbf{\textit{\text{\% If Null Mesh at Position}}} \small
    \State \scriptsize \textbf{\textit{\text{\% If Topometric Occupancy Satisfied}}} \small
        \If{$\beta_\phi \forall \phi \in \{D,C,P,J,I\}$}
            \State $\hat{W} = \hat{W} \cup rand(\hat{M}^\phi)$   \scriptsize \textbf{\textit{\text{\% Add Topometric Unit Mesh}}} \small
            \State \scriptsize \textbf{\textit{\text{\% Recursive Call on Empty Directions}}} \small
            \State $f(W,\gamma(p_{start},\vec{\upsilon}_{start}), p_{start})$
        \EndIf
    \EndIf
\end{algorithmic}
\label{alg:algo2} 
\end{algorithm}
\normalsize 

\begin{figure*}[b!]
    \centering
    \includegraphics[width=\textwidth]{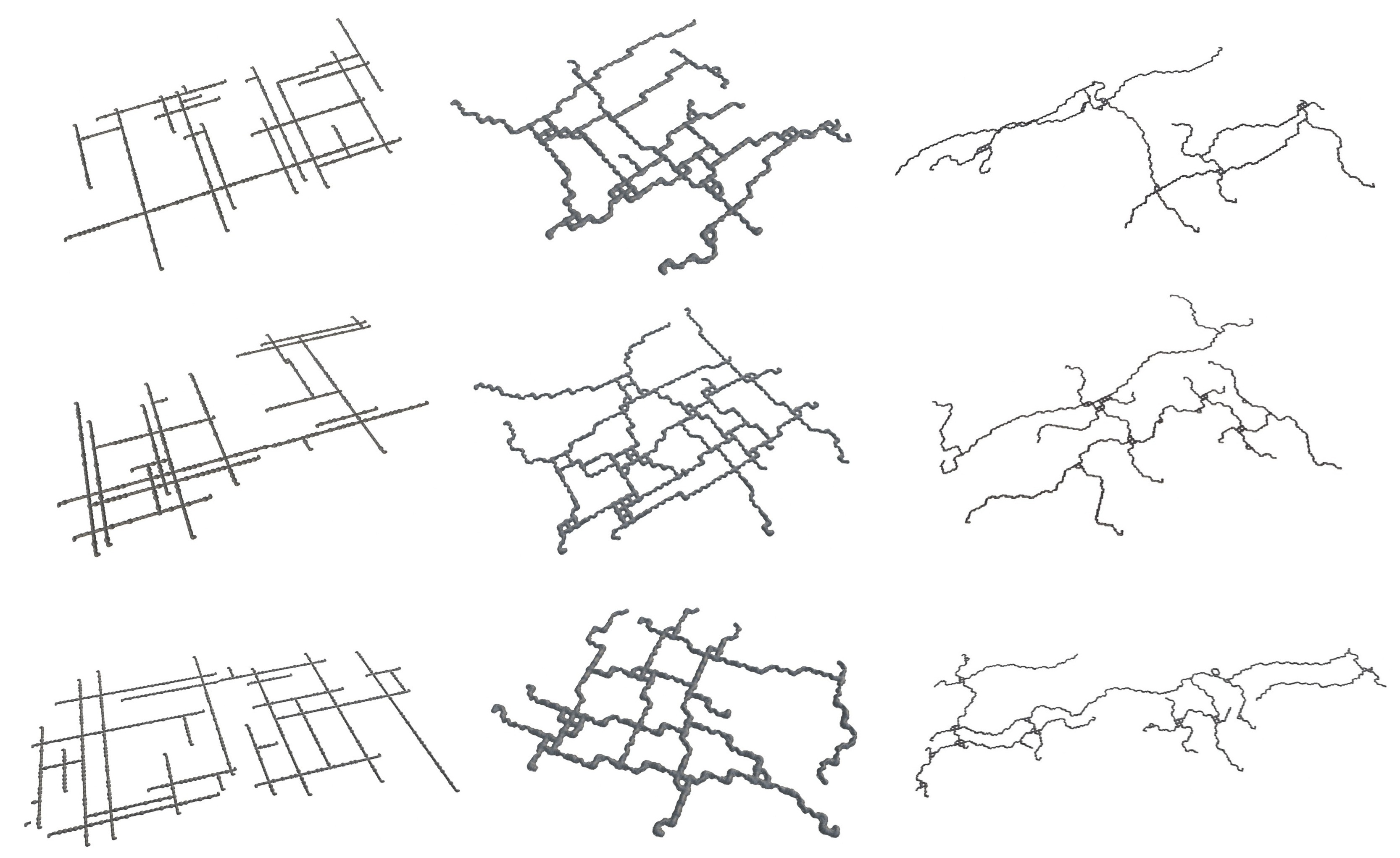}
    \vspace{1mm}
    \begin{minipage}{\textwidth}
        \caption{This figure presents the three subterranean environment types generated by the SubTGraph software: (1) Operational Mines, (2) Natural Caves \& (3) Lava Tubes. The columns of this figure display from left to right topologies with linear, parabolic and sine route descriptors. The parameters utilized to generated such worlds are described in Table \ref{tab:table1}.} \vspace{-3mm}
        \label{fig:figure5}
    \end{minipage}
\end{figure*}

\begin{table*}[b!]
    \centering
    \begin{tabular}{||c|c|c|c|c|c|c|c||}
        \hline
         Environment & Route Description & Min. Length & Max. Width & Levels & Loops & junctions & Intersections\\ \hline\hline
         Operational Mine & Linear & (2-4) km & (80-100) m & 3 & 0 & (0-2) & (0-2) \\ \hline
         Natural Cave & Parabolic & (1-2) km & (40-50) m & 1 & (0-2) & (1-3) & (1-3) \\ \hline
         Lava Tube & Sine & (3-5) km & (40-50) m & 5 & 0 & (1-3) & (0-1) \\ \hline
    \end{tabular} 
    \caption{Parameterization of Underground Environments} \vspace{-3mm}
    \label{tab:table1}
\end{table*}

\section{Validation \& Statistical Analysis} 

\noindent The SubTGraph creation tool is presented as highly variable and controllable to expedite the research of robotic autonomy stacks in simulated subterranean environments. In order to validate this, a benchmark dataset is created utilizing the available parameterization to design three distinct environment types: Operational mines, natural caves and lava tubes. These characteristic environments are generated with different topologies, obstacles and textures. Over the occupancies of the dataset worlds, an analysis of the topological variability in terms of similarity, asymmetry and appearance probability is presented. Moreover, a study on the computational requirements of the creation tool is given to demonstrate its realistic usage by the research community with different available hardware platforms.

\begin{figure*}[t!]
    \centering
    \includegraphics[width=\textwidth]{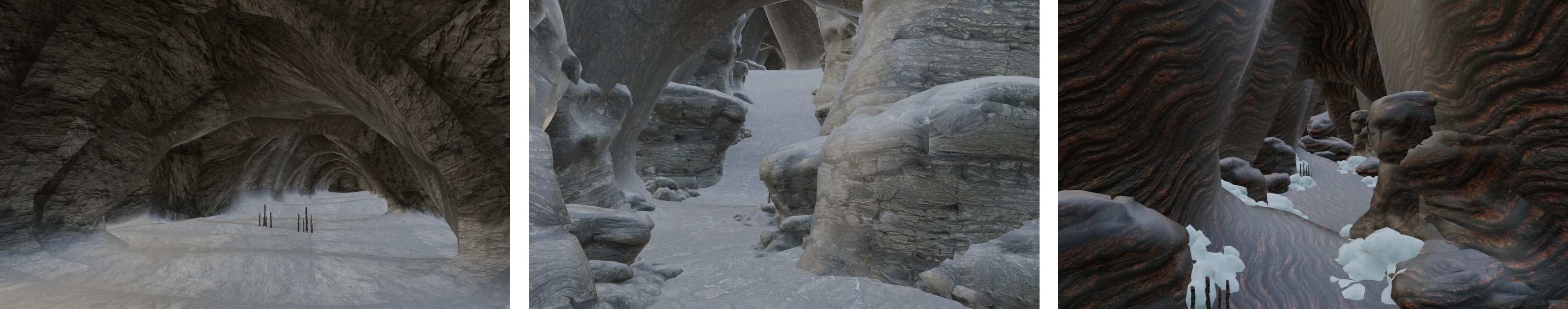}
    \caption{The object spawn and interior textures are controlled by the user specifications. These can be extended to support other types of underground environments. Operational Mines, Natural Caves and Lava Tubes are presented in this figure as three perceptually separate worlds.} \vspace{-2mm}
    \label{fig:figure6}
\end{figure*}

\subsection{Benchmarking Dataset}

\noindent A dataset of 150 world meshes is released with this work, expediting the benchmarking and validation of robotic autonomy algorithms. This dataset is composed of 50 meshes from each environment type according to the parameters of Table \ref{tab:table1}. Figure \ref{fig:figure5} shows example dataset worlds generated with the SubTGraph creation tool. Figure \ref{fig:figure6} displays accordingly the different textured interiors. A set of object tiles from the DARPA world subterranean challenge is used during the mesh generation. These tiles have varying shapes that adequate better to different kinds of environments. In operational mines, we choose wider elongated corridors whereas in lava tubes; wavy sine-like corridors represent better the topology. Natural caves include all  mesh type permutations from within the selected subset of tiles. Nonetheless, users can create their composition of mesh types according to their necessities through a custom configuration file.

Controllability over the textures within the mesh is also possible. Each element is composed of three objects: the cave wall, the rock pile and the striated rock. These objects have an associated texture and can as well be included or excluded during the generation of the subterranean worlds. 
Other parameters such as length and width of the environment are controlled during the post-processing of the generated mesh. A scale ratio is computed between the user-defined requirements and the values obtained from the mesh. Thereafter the mesh is scaled on X,Z axes for width and on Y axes for length. 

\subsection{Topological Analysis}

\begin{figure*}[t!]
    \centering
    \includegraphics[width=\textwidth]{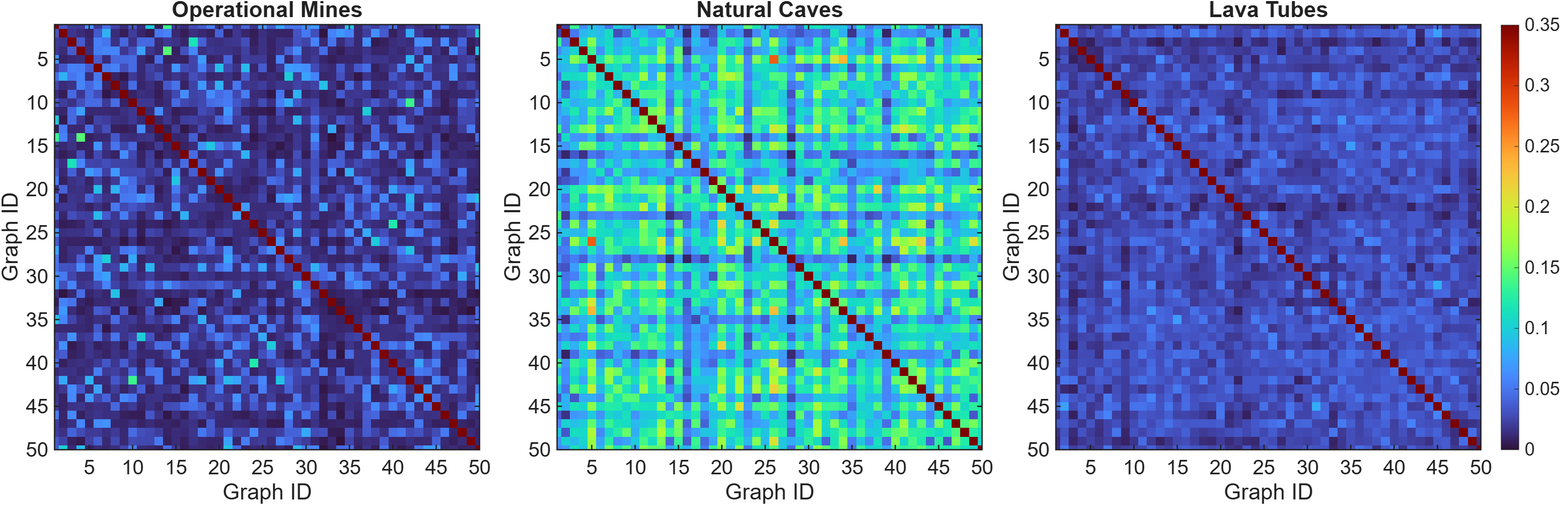} \vspace{-6mm}
    \caption{Similarity Heatmap of environments on linear, parabolic and sine circuits.}
    \label{fig:figure7}
\end{figure*}

\noindent An evaluation of the topological variation is given as Intersection over Union (IoU) in the released benchmark. IoU is measured between two occupancy matrices and provides a metric of topological similarity as per Eq. \ref{eq16}.  This assessment compares each occupancy matrix for every topology type. The objective is to obtain insights on the randomness of the generation process. Overall, the resulting heatmaps as seen in Fig. \ref{fig:figure7} show low similarities. Linear and sine circuits produce the lowest IoU due to their wider dimensions and asset spawn over larger areas. Parabolic circuits on the other hand are instantiated with higher occupancies on the grid, producing increased similarities. \vspace{-3mm}

\begin{equation} \label{eq19}
    \mathrm{IoU}(M_1, M_2) = \frac{\sum_{i,j} \ M_1(i,j)\, M_2(i,j)}{\sum_{i,j} \ \left(M_1(i,j) + M_2(i,j)\right)}
\end{equation}

\begin{figure*}[t!]
    \centering
    \includegraphics[width=\linewidth]{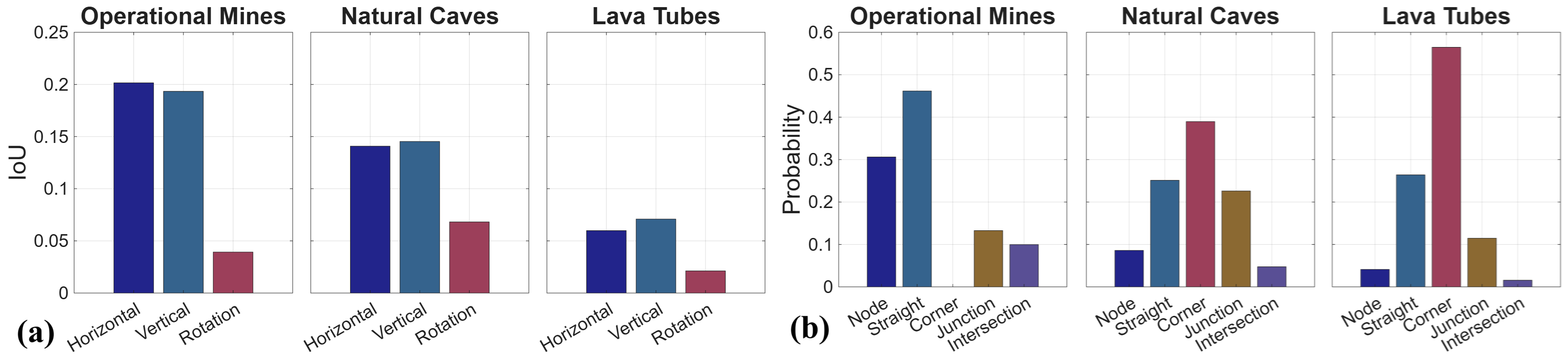} \vspace{-6mm}
    \caption{(a) Symmetry of environments with horizontal, vertical and rotational mirroring. (b) Asset appearance probability distribution on the benchmarked underground environments. Evaluation on linear, parabolic and sine topologies.} \vspace{-3mm}
    \label{fig:figure8}
\end{figure*} 

\subsubsection{Asymmetrical Property}

Asymmetry is a property that many natural patterns satisfy. In this work, we also consider the asymmetry of our environments as a measure of variational richness. The objective is to deduce whether the method can provide natural and unpredictable topologies. Symmetry of environments is hardly measured with a standalone metric. In this evaluation, we have used three values to measure such property: Horizontal, vertical and rotational mirroring. Horizontal and vertical mirroring invert the occupancy matrix on the associated axes as per Eq. \ref{eq17} and Eq. \ref{eq18} respectively. Rotational mirroring on the other hand, rotates the matrix as per equation \ref{eq19}. Each alteration is then compared against the original matrix. Figure \ref{fig:figure8}a shows the evaluation for every topology type on each alteration where none of the examples surpasses the 20\% of similarity. Linear circuits produce the highest similarities as resulting mirrored straight connections maintain their directions. Parabolic and sine circuits present lower similarity by introducing increased topological complexity and therefore more unpredictability within the generation process. \vspace{-3mm}

\small

\begin{equation} \label{eq20}
    \mathrm{\Gamma_\mathcal{H}}(M)= \mathrm{IoU}\bigl(M,\mathcal{H}(M)\bigr); \quad \mathcal{H}(M)(i,j) = M(i, W-j+1)
\end{equation} \vspace{-10mm}

\begin{equation} \label{eq21}
    \mathrm{\Gamma_\mathcal{V}}(M)= \mathrm{IoU}\bigl(M,\mathcal{V}(M)\bigr); \quad  \mathcal{V}(M)(i,j) = M(H-i+1, j)
\end{equation} \vspace{-10mm}

\begin{equation} \label{eq22}
\begin{matrix}
    \mathcal{R}_{\theta}(M) := \text{Matrix } M \text{ rotated by Angle } \theta \\
    \mathrm{\Gamma_\mathcal{R}}(M) = \frac{1}{3}\sum_{n=1}^{3} \mathrm{IoU}\bigl(M,\mathcal{R}_\theta(M)\bigr), \quad \theta = 90n
\end{matrix}
\end{equation}

\normalsize

\subsubsection{Appearance Distribution}

\noindent The probability of appearance of every asset type (i.e. nodes, straight corridors, corners, etc.) is also analyzed to provide an improved insight over the generation randomness. In Figure \ref{fig:figure8}b, the resulting distributions explain the asset usage for each route description. Linear circuits mainly use straight connections. Parabolic circuits utilize less straight connections but higher number of corners and junctions to satisfy wave-like topologies. Sine circuits follow a similar trend with corners dominating the appearance due to increased turns and curves separated over wider areas.

\subsection{Computational Requirements}

\noindent A study of the computation and storage requirements utilized by the SubTGraph creation tool is presented. Different system configurations are chosen to validate the usability under both commercial and academical platforms. Table \ref{tab:table2} shows the chosen configurations ranging from domestic laptops to hardware accelerated computer stations and cluster instances. The creation process utilizes the memory to load the different tiles before composing the world whereas the CPU performs offsetting operations on the vertices. The GPU cannot be utilized as no matrix operations are performed during the mesh generation. 

\begin{table}[b!]
    \centering
    \begin{tabular}{||l|c|c|c||}
        \hline
         \textit{CPU Model} & $N_c$ & \textit{RAM} & \small $GFLOPS_{32}$ \\ \hline\hline
         \small AMD Ryzen 7 4800H & 8 & 16 GB & 268.8 \\ \hline 
         \small AMD Ryzen HX370 & 12 & 32 GB & 489.6 \\ \hline 
         \small Intel Xeon Gold 6338 & 16 & 64 GB & 819.2 \\ \hline 
         \small Intel Core i9-14900K & 24 & 130 GB & 947.2 \\ \hline  
    \end{tabular} \vspace{3mm}
    \caption{System Configurations}
    \label{tab:table2}
\end{table}

\subsubsection{Mesh Generation Time}

\noindent The mesh generation time for every system configuration and environment type is shown in Fig. \ref{fig:figure9}b. From these configurations, the AMD Ryzen HX37 best represents the modern domestic CPUs available to the general public. For this processor, the mesh generation time is on average of 25s for operational mines, 40s for natural caves and 120s for lava tubes. 

\subsubsection{Mesh Storage}

\begin{figure*}[t!]
    \centering
    \includegraphics[width=\linewidth]{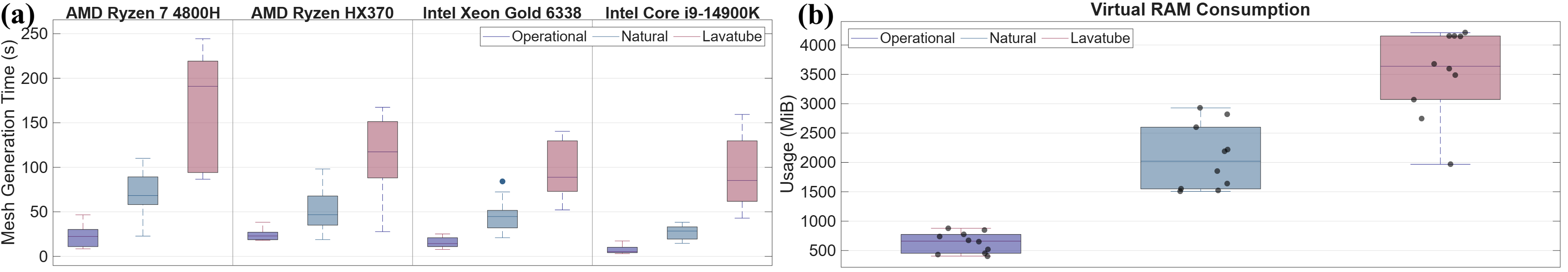} \vspace{-3mm}
    \caption{(a) Mesh generation time on every environment type for system configurations in Table \ref{tab:table2}. (b) Virtual RAM consumption during physics-based simulation.}  \vspace{-3mm}
    \label{fig:figure9}
\end{figure*}

\noindent The mesh sizes in megabytes for each of the 50 benchmark worlds on every environment type are shown in Figure \ref{fig:figure9}a. On average, the operational mines occupy 75MB, the natural caves 400MB and the lavatubes 900MB. However, the SubTGraph tool allows the topology to be saved as an occupancy matrix, storing the world in a few kilobytes. The generation can later be applied over the occupancy matrix to obtain the mesh world. 

\begin{figure}[t!]
    \centering
    \includegraphics[width=\linewidth]{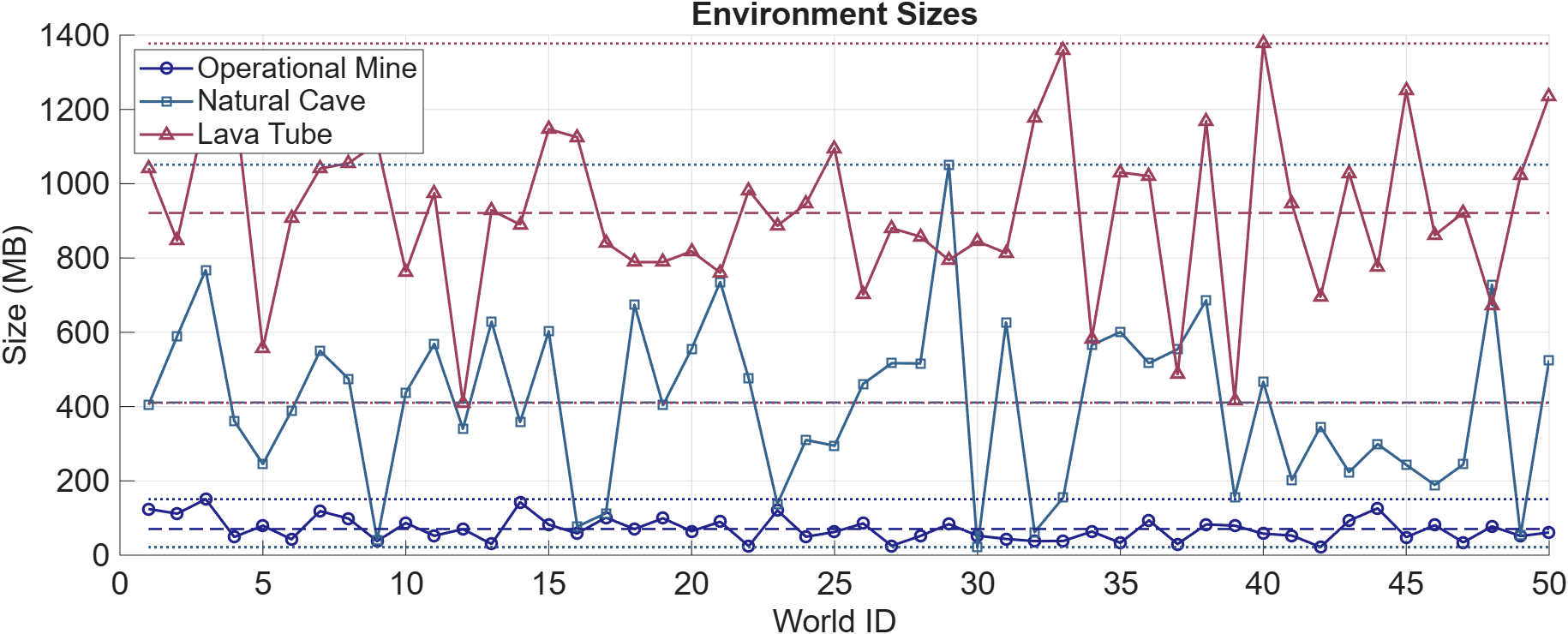} \vspace{-3mm}
    \caption{Environment size distribution on dataset benchmark.}
    \label{fig:figure10}
\end{figure}

\subsubsection{Simulation Memory Consumption}

\noindent These worlds are loaded on the Gazebo physics simulator as seen in Fig. \ref{fig:figure12} for validation of different robotic autonomy algorithms. The interactions of the robotic entity with the world environment are performed through GPU hardware acceleration. A measure of the virtual RAM consumption over a NVIDIA GeForce RTX 4070 GPU during the Gazebo simulations is performed for 10 worlds of each environment type. As seen in Fig. \ref{fig:figure10}, the operational mines occupy on average 700MiB, the natural caves 2GiB and the lavatubes 3.6GiB. 

\section{Application Scenarios}

\noindent The SubTGraph underground world generator is an ideal tool for research of autonomous aerial or terrestrial robots in i.a. mines or caves. The main components of an autonomous framework include path planning, localization, mapping, navigation and safety. In order to showcase the utility, a simulated setup has been made on each environment type. These example cases express the immediate benefits that the topological complexity provides to each of the aforementioned components. 

\subsection{Structural Semantic Segmentation}
\begin{figure}[t!]
  \centering
  \includegraphics[width=0.75\linewidth]{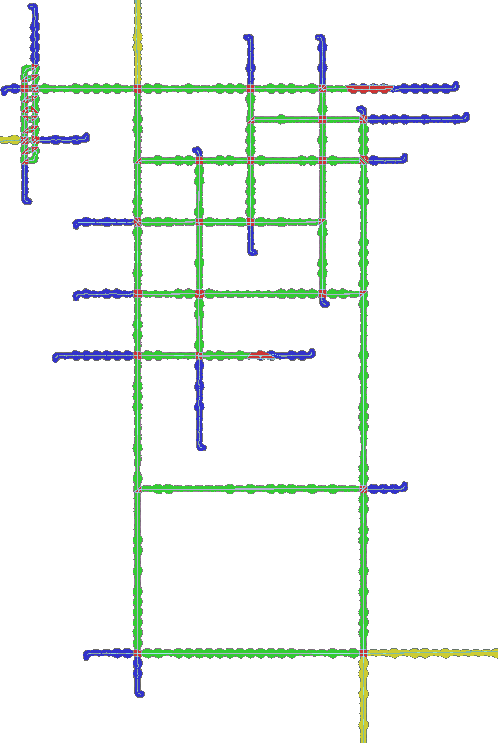} \vspace{-3mm}
  \caption{Topometric map estimated through GRID-FAST over an operational mine. The map is transformed into structural semantic components such as intersections and pathways.}
  \label{fig:figure11}
\end{figure}

\begin{figure}[t!]
  \centering
  \includegraphics[width=\linewidth]{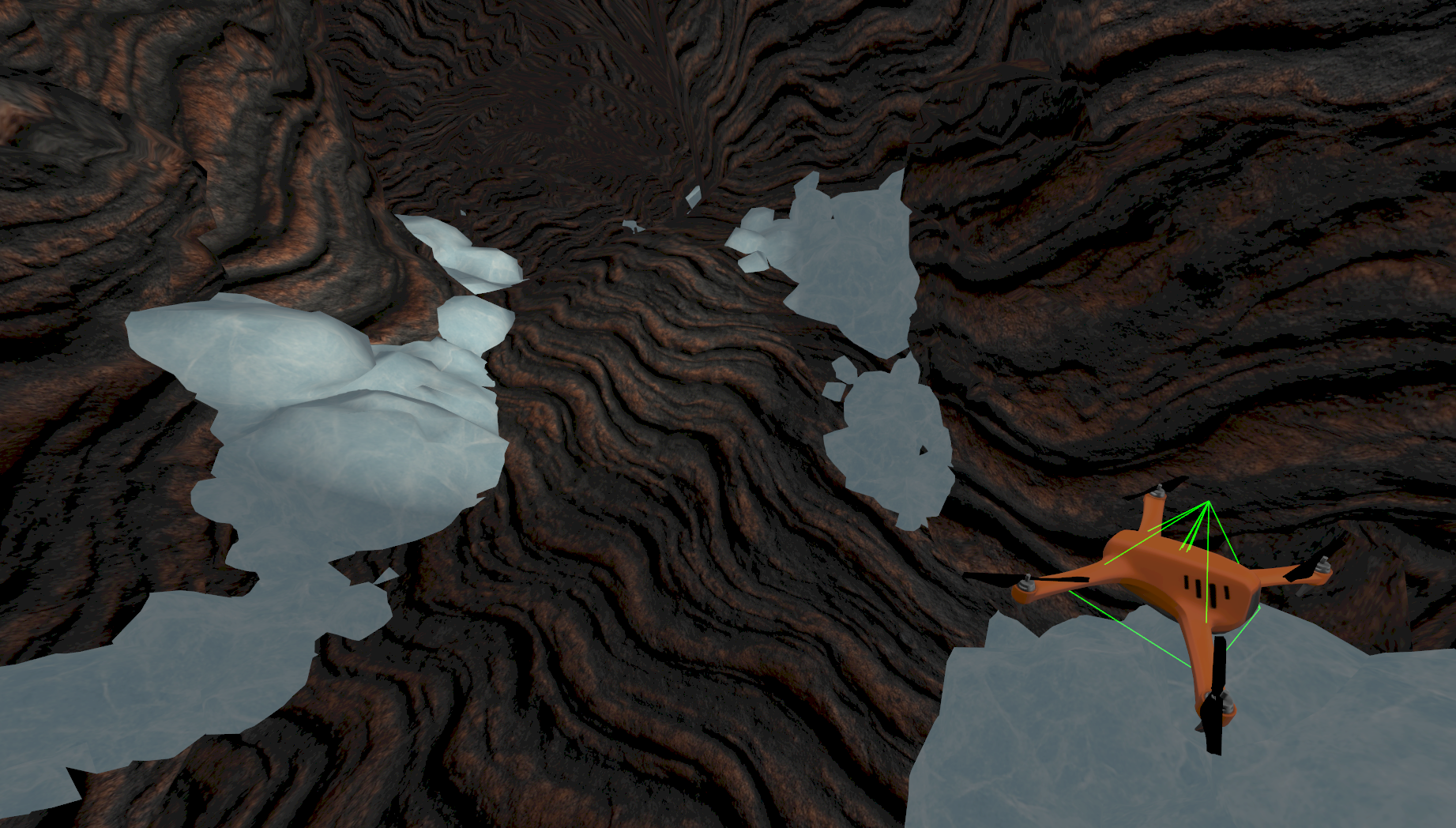}
  \caption{Visualization of the Gazebo physics simulator during validation of an aerial robotic stack on a lava tube environment.}
  \label{fig:figure12}
\end{figure}

\noindent An application example of the SubTGraph creation tool utilizes maps generated from mesh renders to estimate topometric units within an underground environment. GRID-FAST \citep{fredriksson2024grid} performs a topometric map analysis based on 2-dimensional representations to identify underground components such as intersections, dead ends, pathways, and pathways leading to unexplored areas. Figure \ref{fig:figure11} shows the resulting output of GRID-FAST over an operational mine.

The structural semantic topometric map generated by GRID-FAST has a multitude of applications, as the map is reduced to its core structural elements, allowing for a more lightweight representation compared to grid-based or point cloud-based maps. This enables more resource-efficient and faster map operations. The structural semantics can also be utilized for platform reasoning, enabling different behaviors in corridors compared to intersections, as well as semantic objects in the environment that can be represented within the context of their placement in a structural semantic hierarchy. GRID-FAST has already been applied in exploration \citep{Fredriksson2024b} and multi-agent pathfinding \citep{Fredriksson2025}; however, both of these papers use small, highly constrained environments. By utilizing the proposed world generation approach, these works can be extended to operate in larger, more realistic environments and leverage more of the structural information present in the environment.

\begin{figure}[h!]
  \centering
  \includegraphics[width=0.75\linewidth]{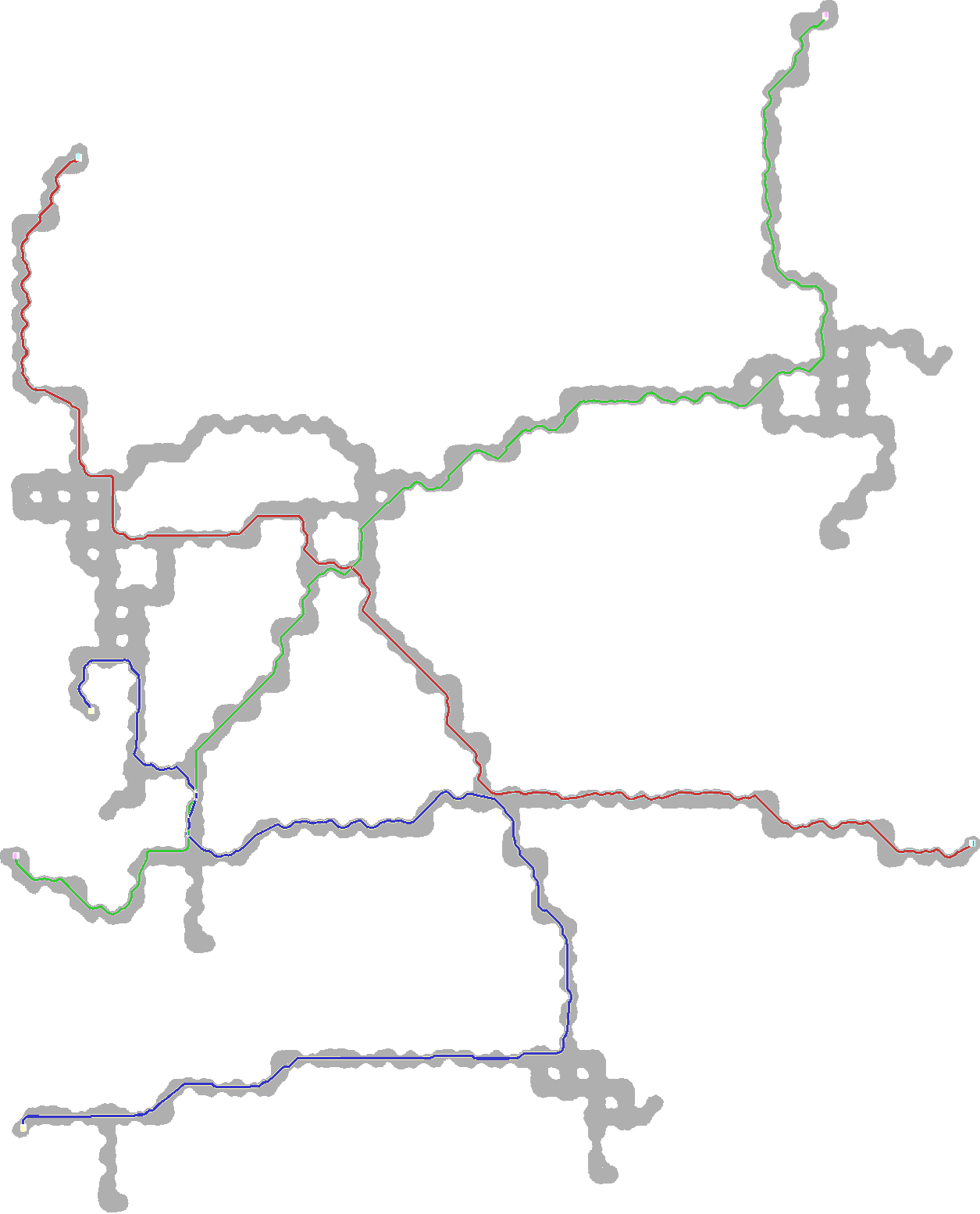} \vspace{-3mm}
  \caption{A$^*_+$T multi-agent path planning over an unstructured natural cave. The method accounts for obstacles and agent path conflicts through temporal estimation.}
  \label{fig:figure13}
\end{figure}

\begin{figure*}[b!]
    \centering
    \includegraphics[width=\linewidth]{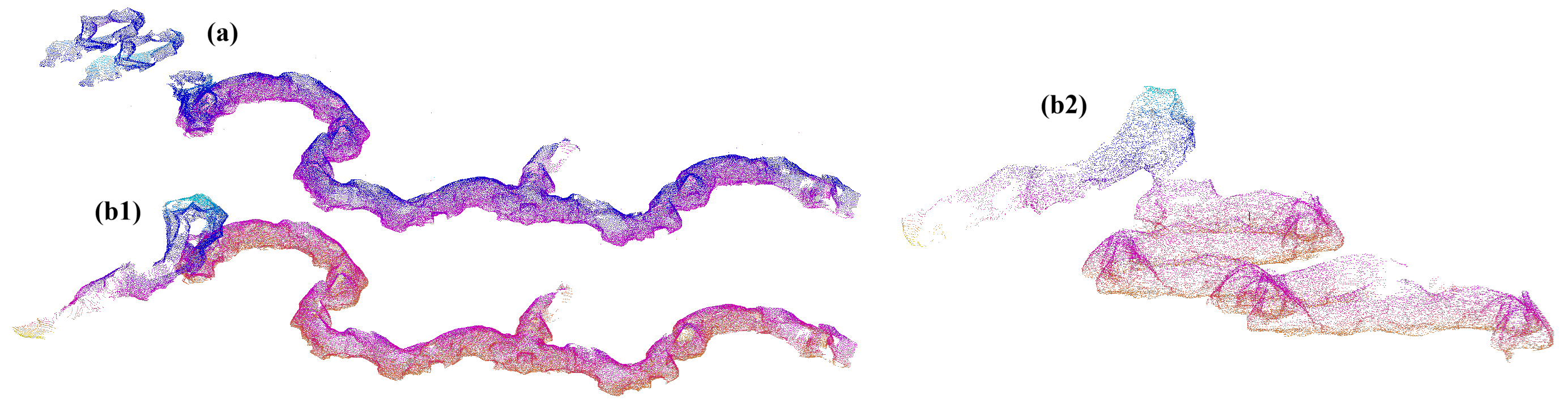} \vspace{-3mm}
    \caption{Evaluation of (a) FastLIO and (b) DLIO during an aerial exploration mission for the case of vertical shaft navigation. } \vspace{-3mm}
    \label{fig:figure14}
\end{figure*} 

\subsection{Multi-robot Path Planning}
\noindent Another application of the SubTGraph creation tool is multi-robot path planning. This study employs the multi-agent path planner A$^*_+$T \citep{nordstrom2025time} to estimate a global path planning utilizing established maps of natural cave environments. A$^*_+$T operates as a distributed multi-agent path planning system, enabling each agent to compute an optimal path based on a risk heuristic \citep{KARLSSON2023119408} and the paths broadcasted by other agents. The incorporation of these paths enhances the risk assessment, accounting for the temporal and spatial context of potential overlaps. 

A common challenge faced by traditional path planners is their reliance on binary traversability, which can lead to the inflation of obstacles to maintain safety margins. This inflation can obstruct narrow passages that fall below the requisite safety margin, while simultaneously allowing for unnecessarily expansive margins in more open areas. In contrast, the heuristic employed by A$^*_+$T preserves appropriate safety margins in wider passages while permitting traversal through narrower spaces, provided that a critical safety margin is maintained. Figure \ref{fig:figure13} illustrates a static analysis of the paths generated by A$^*_+$T within various natural cave environments, where three different agents have been assigned e.g. an exploration path given their initial and objective positions. 

However, the temporal dynamics of multi-agent interactions present additional complexities that are less visually apparent. Many tests reveal whether paths run parallel or intersect. Since each agent occupies its designated path at any singular moment, the remainder of its trajectory remains available for occupation by other agents. This results in overlapping and intersecting paths when examined through a static lens. For a comprehensive evaluation of multi-agent routing efficiency, it is essential to consider the temporal aspects of agent movement and interaction. 

\begin{figure}[t!]
    \centering
    \includegraphics[width=\linewidth]{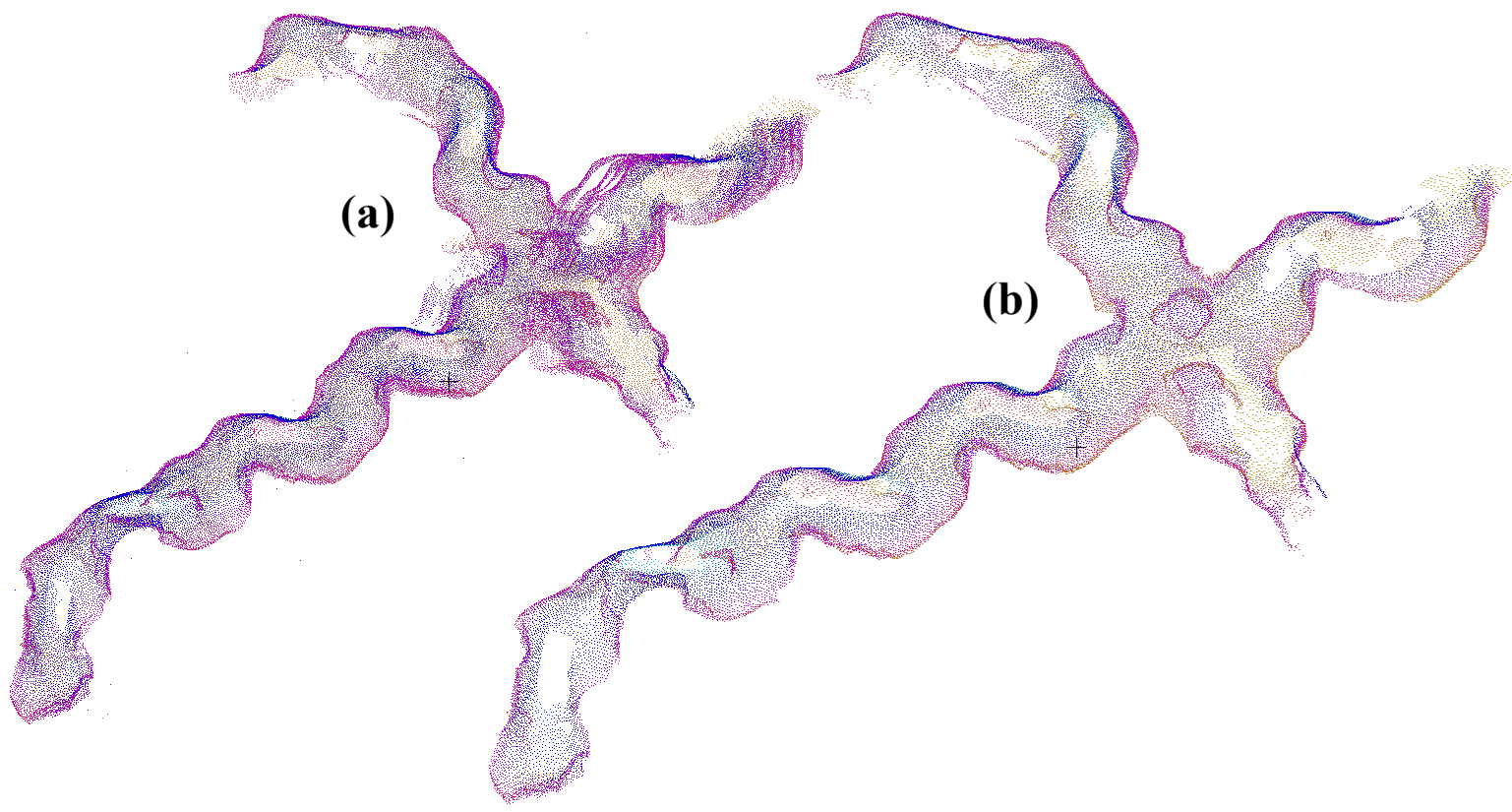} \vspace{-3mm}
    \caption{Evaluation of (a) FastLIO and (b) DLIO during an aerial exploration mission for the case of backtracking.}
    \label{fig:figure15}
\end{figure}

\subsection{Simultaneous Localization \& Mapping}
\noindent The Gazebo physics simulator is utilized to validate an aerial robotic algorithm stack within a lava tube environment. In this application, the navigation of the UAV is teleoperated by utilizing a remote controller while the localization and mapping is performed by two distinctive LIO implementations. This basic algorithm stack allows the validation of an essential component during autonomous missions in underground environments. Several corner case subterranean sections are selected to evaluate the performance of FastLIO \citep{xu2022fast} and DLIO \citep{chen2022direct}. A 3-dimensional map is generated based on the estimated odometry and the simulated lidar scans by identifying unique features in the point cloud. These features are merged with the IMU information to localize the robot in real time. 

The first corner case targets a SLAM evaluation during the navigation of a vertical shaft. The odometry estimation of FastLIO fails when navigating towards a different level, as seen in Fig. \ref{fig:figure14}a. The estimation performed by DLIO does not completely drift but the method fails in reconstructing the geometry of the shaft, as seen in Fig. \ref{fig:figure14}b1,b2. The mapping of the pathway previous to the shaft is correctly constructed, suggesting that both LIO methods have not been stress tested for vertical navigation on such cluttered environment. Figure \ref{fig:figure15} displays another corner case in which the robotic platform reaches a dead end and backtracks to the starting intersection area. The results follow a similar trend where FastLIO drifts when reaching the starting point, as seen in Fig. \ref{fig:figure15}a; while DLIO correctly maintains the odometry and mapping, as seen in Fig. \ref{fig:figure15}b. This analysis allows researchers to adequately select an algorithm suiting experimental requirements before field deployment.

\section{Discussion}

\noindent The SubTGraph world creation tool provides a platform to generate underground topologies according to user specifications. This tool is ideal for testing methods that require autonomy in complex environments, e.g. odometry estimation in loop closures or navigation in corner-junction sections. When comparing the cost of testing such niche and specific problems in field tests with respect to simulated environments, the benefits of the simulation are higher. Arranging field tests requires i.a. an analysis of the existing underground sites, a lookup of the specific regions of interest to tackle the highlighted problem, the request of permits to use robots in the environment and the transportation costs. This work considers that field tests are without doubt the final step towards validating any method. However, an intermediate step is necessary to tackle the semantical errors together with the corner cases before the final stages of field testing. In this way, both time and explainability are optimized during real validation. Moreover, the availability of hundreds of generated underground worlds allows the possibility to benchmark and statistically validate the generalization of any method.

World generators show great possibilities for expediting the simulation efforts of underground robotic solutions. figures, point clouds and robot states can be extracted during simulation, increasing data availability for research fields such as Deep Learning. Limitations of SubTGraph include the effects of asset repeatability and environment discretizations, i.e. methods that rely on feature identification may break when a repeated tile appears at different positions. Moreover, this tool does not provide the fidelity of a digital twin; rather the simulated means to benchmark a method on extensive topological varieties. 

\section{Conclusion} 
\noindent In this work we present SubTGraph, an underground world generator with topological, geometrical and textural controllability that supports the research efforts of underground robotic solutions. An application scenario of different components in an autonomous algorithm stack showcases the usage of such tool for statistical validation. With this work, an open-sourced dataset of 150 worlds with respectively operational mines, natural caves and lava tubes is released. 
The future work shall focus on investigating 3-dimensional geometrical transformations of the tile components to advance towards continuous underground representations that include randomized textures for feature detection and obstacle location. 

\bibliographystyle{unsrt} 
\bibliography{references}

\onecolumn
\appendix

\clearpage
\section{Method Validation on SubTGraph Benchmark Dataset}
\subsection{GRID-FAST Topometric Validation}
\begin{figure}[h!]
    \centering
    \includegraphics[width=\textwidth]{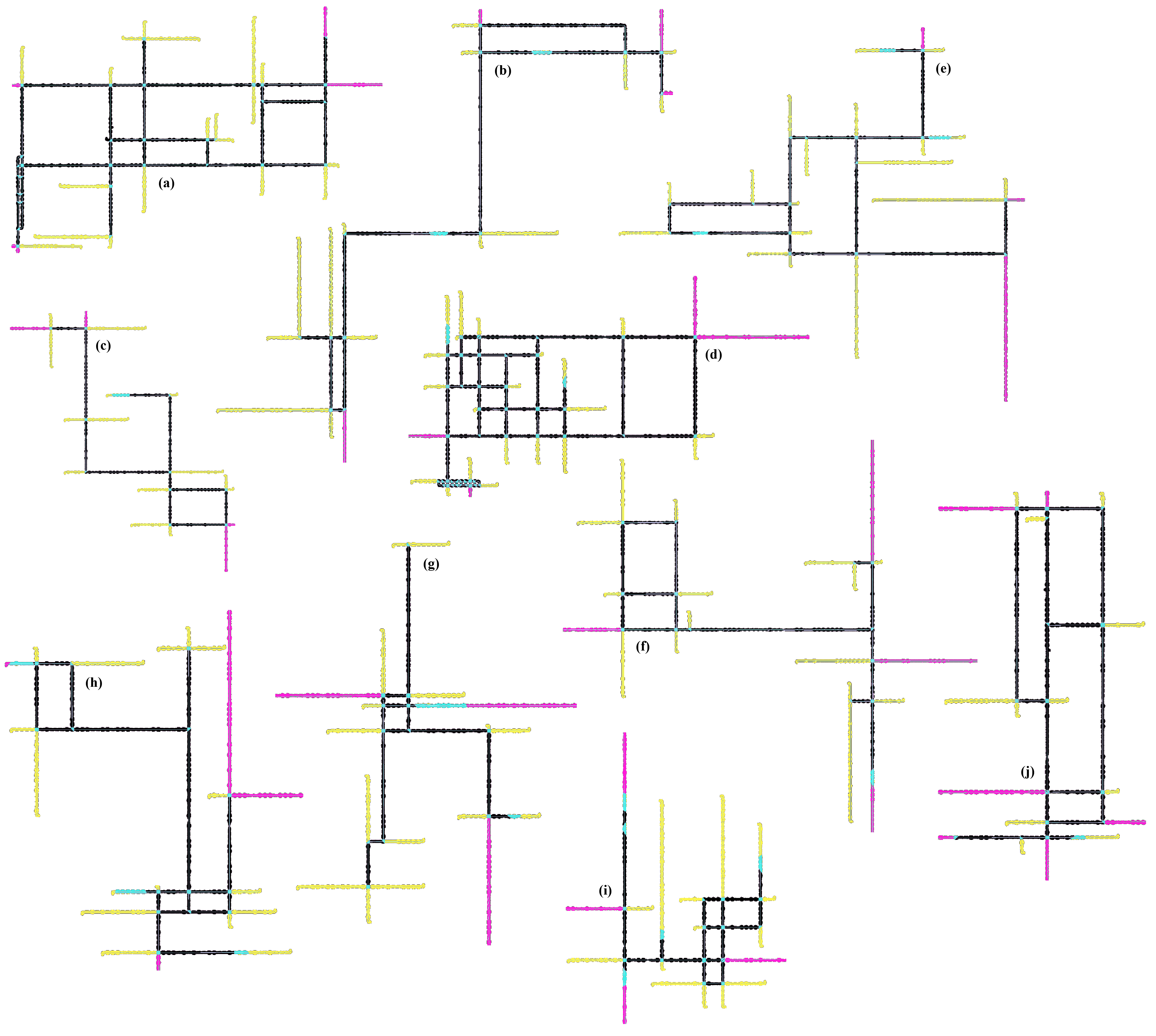}
    \caption{GRID-FAST validation on dataset subsection. Black color indicates straight areas, yellow sections indicate dead ends, blue sections indicate intersections and pink sections indicate frontiers. The results display failed identification of intersections (b),(h),(g).}
    \label{fig:appendix1}
\end{figure}

\clearpage
\subsection{ASPT Planning Validation}
\begin{figure}[h!]
    \centering
    \includegraphics[width=\textwidth]{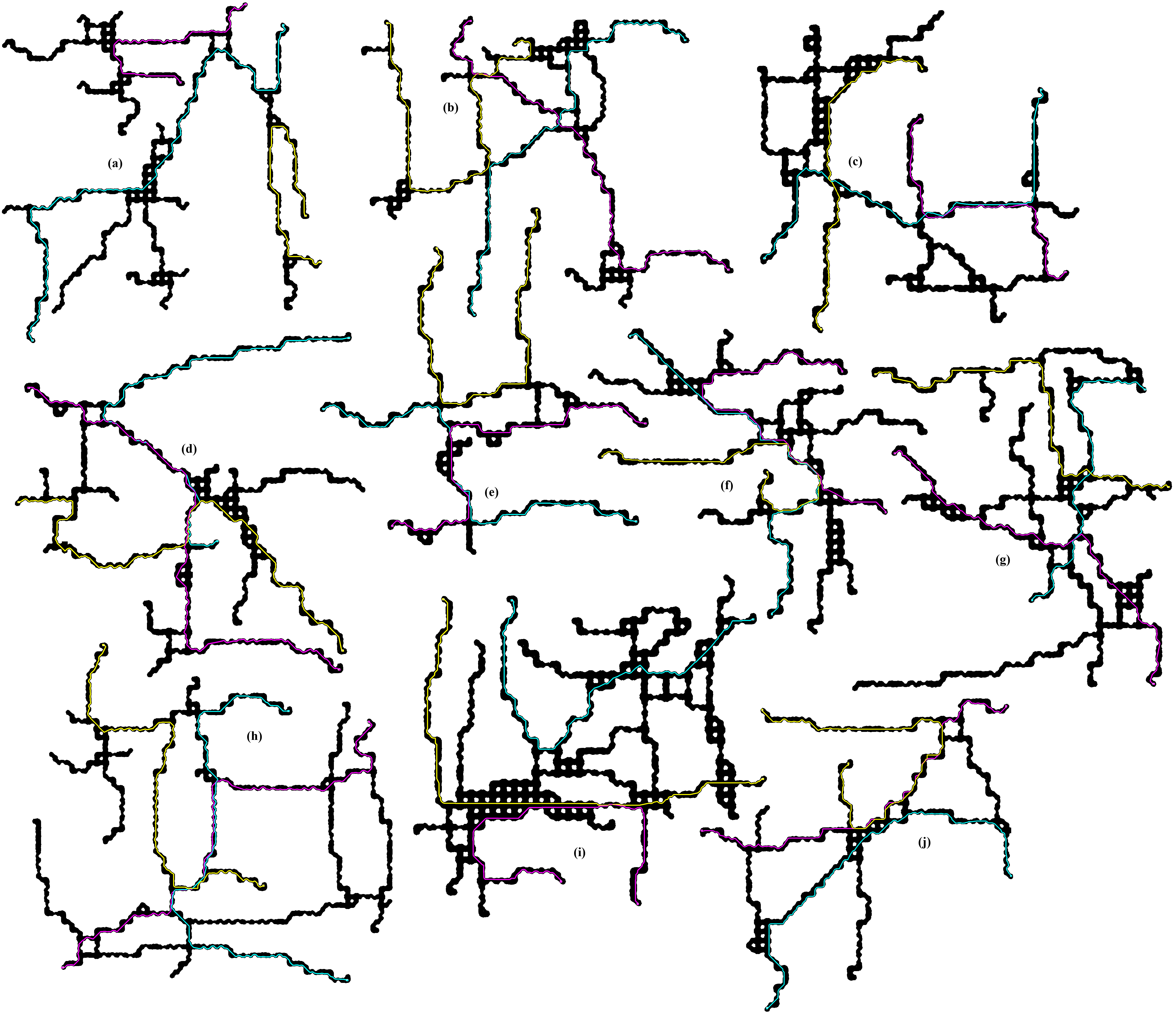}
    \caption{ A$^*_+$T validation on dataset subsection. Three agents are spawned with an associated objective position on different sections of the natural cave. The optimization process searches for the fastest route while evading other agents' paths. Results on (d),(e),(j) display overlapping paths, showing  A$^*_+$T's capability to utilize the same space for multiple agents at different times.}
    \label{fig:appendix2}
\end{figure}

\clearpage
\subsection{Localization \& Mapping Validation}
\begin{figure}[h!]
    \centering
    \includegraphics[width=\textwidth]{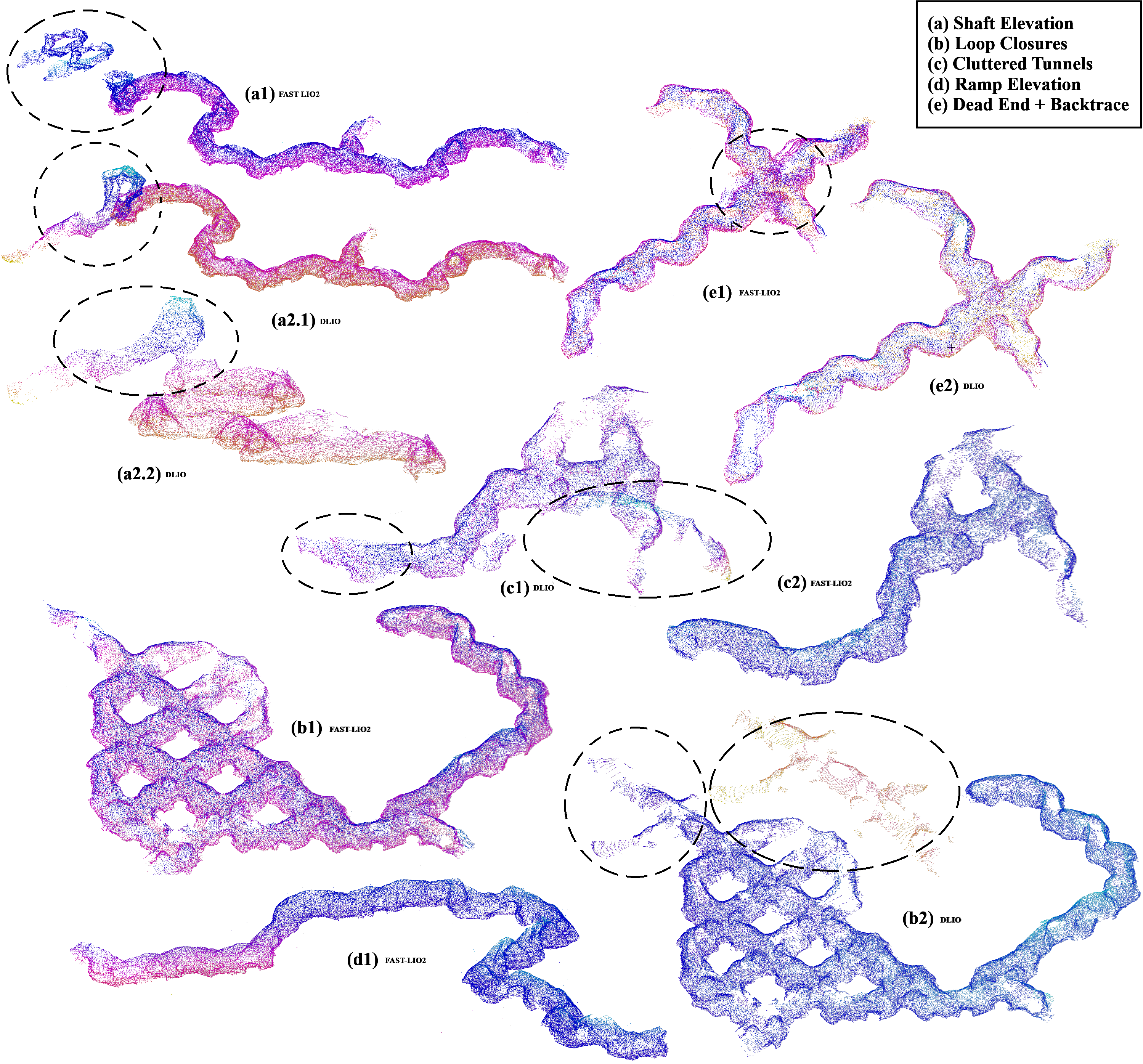}
    \caption{LIO algorithm validation. Benchmarking on different sections of the generated underground worlds, two SOTA methods are executed to identify cases in which the estimated odometry drifts or the mapping breaks. Five cases are identified as susceptible to failure during underground missions. For each, the resulting map is presented with highlighted failures.}
    \label{fig:appendix3}
\end{figure}

\end{document}